\documentclass{article}

\usepackage{arxiv}

\usepackage[utf8]{inputenc} 
\usepackage[T1]{fontenc}    
\usepackage{hyperref}       
\usepackage{url}            
\usepackage{booktabs}       
\usepackage{amsfonts}       
\usepackage{nicefrac}       
\usepackage{microtype}      
\usepackage{lipsum}
\usepackage{graphicx}

\usepackage{booktabs}       
\usepackage{tabularx}       
\usepackage{multirow}       
\usepackage{array}          
\usepackage{multicol}
\usepackage{siunitx}  
\usepackage{enumitem}
\usepackage{bm, booktabs, amsmath}

\usepackage[capitalize]{cleveref}
\crefname{section}{Sec.}{Secs.}
\Crefname{section}{Section}{Sections}
\Crefname{table}{Table}{Tables}
\crefname{table}{Tab.}{Tabs.}

\title{Mono3R: Exploiting Monocular Cues for Geometric 3D Reconstruction}

\author{
  Wenyu Li, Sidun Liu, Peng Qiao and Yong Dou \\
  National University of Defence Technology \\
  Changsha, China \\
  \texttt{\{wenyu18, liusidun, pengqiao, yongdou\}@nudt.edu.cn} \\
}

\begin{document}
\maketitle
\begin{abstract}
Recent advances in data-driven geometric multi-view 3D reconstruction foundation models (e.g., DUSt3R) have shown remarkable performance across various 3D vision tasks, facilitated by the release of large-scale, high-quality 3D datasets. However, as we observed, constrained by their matching-based principles, the reconstruction quality of existing models suffers significant degradation in challenging regions with limited matching cues, particularly in weakly textured areas and low-light conditions. To mitigate these limitations, we propose to harness the inherent robustness of monocular geometry estimation to compensate for the inherent shortcomings of matching-based methods. Specifically, we introduce a monocular-guided refinement module that integrates monocular geometric priors into multi-view reconstruction frameworks. This integration substantially enhances the robustness of multi-view reconstruction systems, leading to high-quality feed-forward reconstructions. Comprehensive experiments across multiple benchmarks demonstrate that our method achieves substantial improvements in both mutli-view camera pose estimation and point cloud accuracy.
\end{abstract}


\begin{figure}[t!]
  \centering
  \includegraphics[width=\textwidth]{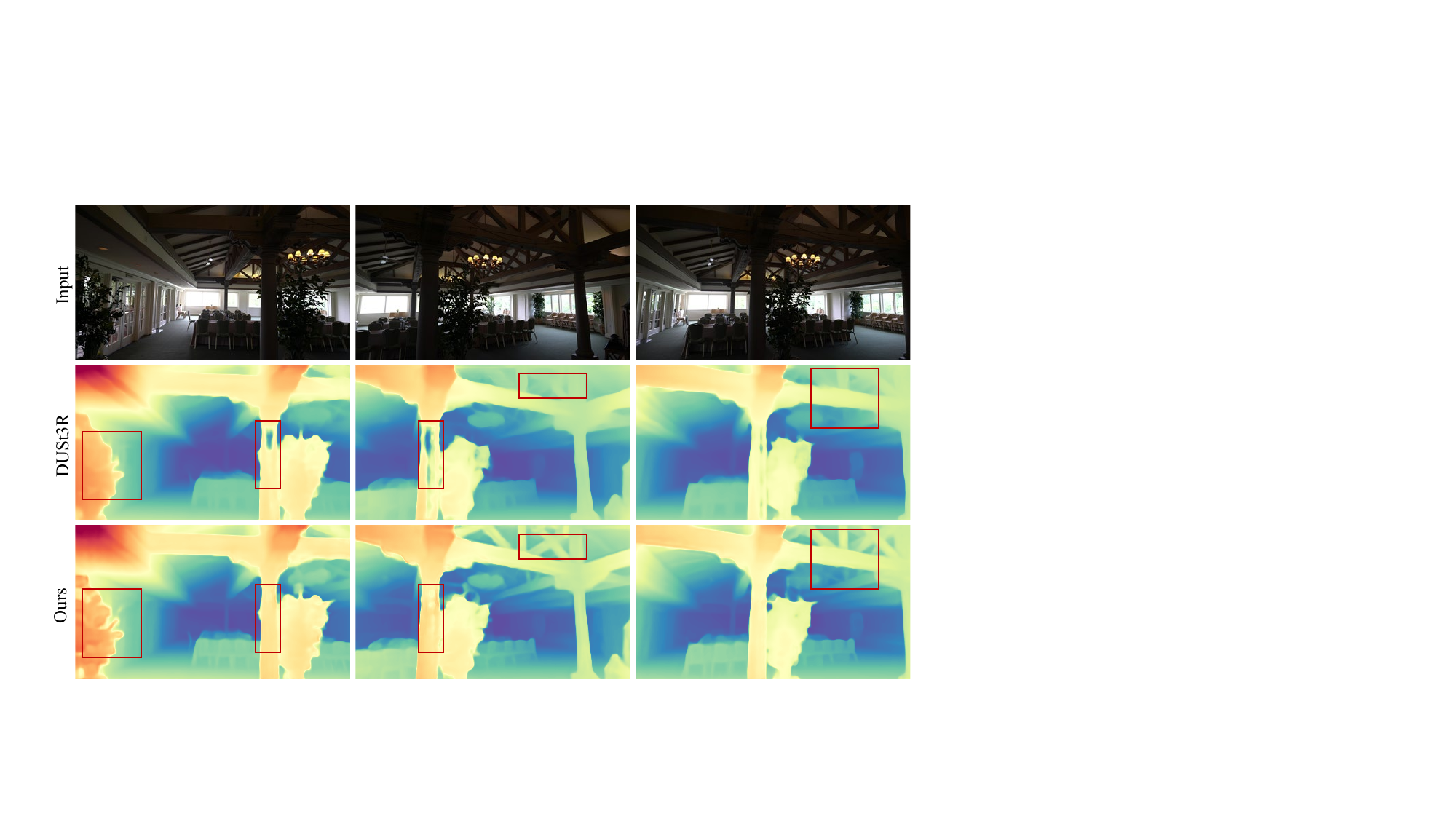}
  \caption{
  In this paper, we reveal the limitations of DUSt3R in reconstructing textureless regions and fine structures, as demonstrated in the 2nd row. This fundamentally stems from matching-based approaches, where matching consistency proves difficult to maintain in such challenging areas. To address these limitations, we propose Mono3R, which integrate the robustness of monocular geometry estimation into DUSt3R. Our method can both reconstruct accurate geometry in textureless regions and recover fine structural details, as shown in 3rd row.
  }
  \label{fig:teaser}
\end{figure}

\section{Introduction}

The recovery of dense geometric structures from 2D imagery represents a cornerstone challenge in computer vision research, with a rich history spanning several decades\cite{xukai1, xukai2, tsai84uniqueness, triggs99camera, triggs96factorization, tuceryan98texture, tuytelaars99image}.
 This capability enables applications across diverse domains including robotic navigation\cite{silpa-anan04localization, geiger13vision, ramezani22aeros:, stumberg22dm-vio:}, augmented reality systems\cite{smith99improving, azuma97a-survey}, autonomous vehicle perception\cite{brooks91elephants, behl17bounding, chitta21neat:, geiger12are-we-ready}, and medical diagnostics\cite{shao22self-supervised, burner11texture}. 
The intrinsic ill-posed nature of inferring 3D geometry from 2D projections has spawned multiple specialized research directions, each addressing distinct aspects of the reconstruction pipeline: photogrammetric techniques like Structure-from-Motion (SfM)\cite{schoenberger2016sfm, cui2017hsfm, wei2020deepsfm, wei20deepsfm:, wang24vggsfm:}, optimization frameworks including Bundle Adjustment (BA)\cite{teo10bundle, triggs00bundle, wen23bundlesdf:, romberg13bundle}, dense depth estimation methods such as Multi-View Stereo (MVS)\cite{schonberger16pixelwise, mvsnet} and real-time systems exemplified by SLAM implementations\cite{slama80manual, thrun02slam, yan24gs-slam:}.
The traditional paradigm for 3D geometric reconstruction relies on elaborate, multi-stage processing pipelines, which demand significant effort for effective integration.

The field has witnessed a paradigm shift toward an approach previously deemed infeasible --- directly regressing the pointmap, from a pair of uncalibrated images without prior scene information. 
Trained on millions of image pairs with ground-truth annotations for depth and camera parameters, the representative method, DUSt3R\cite{wang24dust3r:}, shows remarkable performance and cross-domain generalization across various real-world scenarios.
While demonstrating impressive results, DUSt3R and its variants~\cite{yang2025fast3r, cut3r, mast3r, duisterhof24mast3r-sfm:, wang2024spann3r} fundamentally derive pointmap from patchwise similarity matching.
The underlying architecture of these methods implicitly assumes visible correspondences in both images.
However, matching-based principle poses challenges in ill-posed regions with limited matching cues, e.g., occlusions, textureless areas and repetitive/thin structures, as shown in \cref{fig:teaser}.

While multi-view reconstruction suffers from inherent matching limitations, monocular 3D estimation has evolved as a parallel approach that circumvents mismatching artifacts.
The field has witnessed decades of advancement\cite{agarwal06recovering, ranftl16dense, piccinelli24unidepth:, saxena23the-surprising}. 
Recent breakthrough, MoGe\cite{wang24moge:}, has introduced a novel direct monocular 3D pointmap estimation method and demonstrates significant progress in terms of prediction quality and generalization to new scenes. 
Nevertheless, when applying to multi-view scenarios, monocular methods exhibit their own characteristic constraints: the predictions for multiple images typically lack consistency.
This analysis motivates an intriguing research question: \textit{Can we marry the robust geometric priors of monocular geometry estimators with the multi-view consistency of multi-view matching systems?}
The complementary nature of these approaches - leveraging monocular robustness alongside multi-view precision and consistency - presents a promising avenue for overcoming current limitations in 3D reconstruction.

Based on these insights, we propose Mono3R, a novel approach that fully combines the complementary strengths of monocular and stereo algorithms and overcomes the limitations from the lack of matching cues. 
Given training pair view images, we first separately infer geometry for each image with pre-trained monocular and pairwise geometry estimation models, then we register the monocular point clouds into a unified coordinate system with pairwise pointmap. 
Subsequently, we refine the pairwise pointmap through the refinement module guided by monocular pointmaps and features, which not only maintains the consistency of multi-view point maps but also leads to more robust results in situations that are hard to match (e.g., occlusions, texture-less regions and reflective surfaces). Thanks to the monocular-guided refinement module, the quality of dense geometry reconstruction is enhanced (See \cref{fig:teaser}).  

We conduct comprehensive experiments across five benchmark datasets: object-level DTU~\cite{dtudataset}, indoor scenes (7Scenes~\cite{7scenes} and Neural-RGBD~\cite{sturm2012benchmark}), and unbounded environments (ETH3D~\cite{eth3d} and Tanks \& Temples~\cite{tnt}) to evaluate both multi-view camera pose estimation and point cloud reconstruction accuracy.
Under various evaluation settings, our Mono3R consistently outperforms DUSt3R~\cite{wang24dust3r:} and its subsequent variants Spann3R\cite{wang2024spann3r} and Fast3R\cite{yang2025fast3r}. Most notably, for indoor scenarios, Mono3R achieves a \textbf{13\%} improvement in pose estimation accuracy ($\text{MAA}_{30}$) over DUSt3R, demonstrating the effectiveness of jointly optimizing multi-view reconstruction with monocular geometry estimation.

Our main contributions can be summarized as follows:
\begin{itemize}
\item We identify shortcomings of both monocular and multi-view methods and introduce Mono3R, an approach that fully leverages the monocular geometry and feature priors to help estimate multi-view geometry. Our monocular-guided refinement module demonstrates effectiveness, particularly in overcoming matching-based failures.
\item Through extensive experiments, Mono3R achieves promising results on several downstream tasks, including camera pose estimation and pointmap reconstruction. In particular, Mono3R offers key advantages and robustness over prior works in challenging areas; e.g., occlusions, textureless areas.
\end{itemize}

\begin{figure*}[t!]
  \centering
  \includegraphics[width=\textwidth]{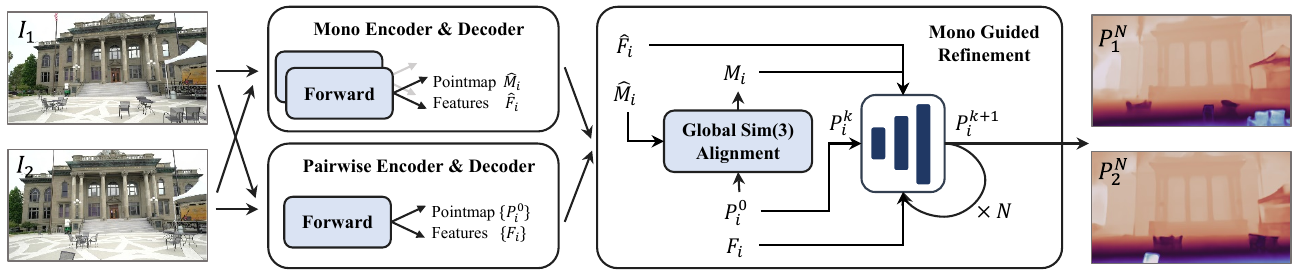}
  \caption{Our framework consists of two complementary branches and a refinement module. 
  The pairwise branch processes image pairs through feature matching to simultaneously extract cross-image feature correspondences and regress 3D point clouds
  The monocular branch processes individual images to extract view-specific geometric information.
  The mono-guided refinement module first performs global Sim(3) alignment to establish a unified coordinate system for the monocular outputs, then iteratively optimizes the pairwise reconstruction for improved accuracy.
}
  \label{fig:method}
\end{figure*}

\section{Related Works}

\paragraph{Traditional Geometric 3D Reconstruction}
Geometric 3D Reconstruction is a core computer vision task that recovers the 3D structure of a scene from multiple 2D images. It involves estimating camera parameters (intrinsic and extrinsic) and reconstructing the scene as a point cloud or mesh.
Traditional methods for geometric 3D reconstruction are broadly categorized into Structure from Motion (SfM)\cite{schonberger16structure-from-motion} and Multi-View Stereo (MVS)\cite{schoenberger2016mvs}.
SfM starts by extracting and matching features\cite{liu08sift, lowe07sift, tsay12sift}, triangulating them into 3D points, and refining the results via bundle adjustment\cite{tang2018ba, agarwal2010bundle}. 
Among existing tools, COLMAP~\cite{schoenberger2016sfm} has emerged as the most widely used framework due to its robustness and scalability.
Once camera parameters are estimated, Multi-View Stereo (MVS) takes over to generate dense reconstructions.
MVS leverages photometric consistency or geometric constraints (e.g., plane-sweep stereo\cite{baillard00a-plane-sweep}, patch-based matching\cite{barnes09patchmatch, furukawa10towards}) to estimate depth maps for each viewpoint. These depth maps are then fused into a unified dense point cloud or mesh.
Despite their effectiveness, traditional geometric reconstruction pipelines face several limitations:
they often require lengthy offline optimization, are sensitive to textureless regions or repetitive patterns, and may struggle with large-scale scenes due to computational complexity. 

\paragraph{Learning-based Geometric 3D Reconstruction}
In the early stages of learning-based geometric 3D reconstruction, the approach to applying deep learning techniques primarily involved using them to replace specific components within traditional 3D reconstruction pipelines, rather than implementing an end-to-end substitution, i.e., feature extraction\cite{daniel17superpoint:, yi16lift:}, matching\cite{sarlin20superglue:, sun2021loftr, chen2022aspanformer}, BA\cite{tang2018ba}, multi-view depth estimation\cite{mvsnet, geomvsnet}, etc.
Although integrating deep learning by replacing individual components allows its incorporation into 3D reconstruction frameworks, the inability to achieve an end-to-end process ultimately restricts further improvements in 3D reconstruction performance and usually involves a sequential structure vulnerable to noise in each subtask.
Most recently, the revolutionary approach, DUSt3R\cite{wang24dust3r:}, directly estimates aligned dense point clouds from a pair of views, similar to MVS but bypassing estimation of camera parameters and poses. It unifies all subtasks by directly learning to map an image pair to 3D, followed by an optimization-based global alignment to bring all image pairs into a common coordinate system.
Several extensions of DUSt3R have been proposed to address different aspects. 
MASt3R\cite{mast3r} enhances local feature matching through the introduction of a dedicated feature head. This additional module further improves the accuracy of point maps, thereby validating the effectiveness of the multi-task learning for geometric 3D reconstruction.
MonST3R\cite{zhang24monst3r:} explores dynamic scene reconstruction through a data-driven approach.
To bypass post-optimization, Spann3R\cite{wang2024spann3r} processes images sequentially (e.g., from video) and incrementally reconstructs scenes using a sliding-window network with a learned spatial memory module.
Fast3R\cite{yang2025fast3r} eliminates sequential dependencies, effectively generalizing DUSt3R\cite{wang24dust3r:} to arbitrary view configurations.
In this work, we take a step further to improve the quality of the reconstruction by exploiting monocular cues.

\paragraph{3D reconstruction from a single image}
Several approaches have been developed for reconstructing specific object categories from single images, such as human bodies~\cite{saito19pifu:}. Another research direction, referred to as \textit{monocular geometry estimation}, focuses on recovering general 3D scene structures from individual images.
The standard pipeline typically involves two sequential steps: (1) depth prediction~\cite{depth_anything_v2, ocal20realmonodepth:, li2018megadepth} followed by (2) camera intrinsic estimation~\cite{deutscher2002automatic, hold2018perceptual}. These predicted components can be combined through projection equations to generate point cloud representations. For example, LeReS~\cite{leres} introduced a two-stage approach featuring affine-invariant depth prediction with a subsequent module for recovering scene shift and camera focal length. UniDepth~\cite{piccinelli24unidepth:} proposed a novel self-promptable camera module that predicts dense camera representations to condition the depth estimation.
In contrast to previous methods, MoGe~\cite{wang24moge:} represents a significant advancement by directly predicting the pointmap from single images. This approach combines large-scale multi-domain training data with carefully designed affine-invariant point map representations and training objectives, enabling more effective geometry learning. In this work, we leverage recent advances and enhanced robustness in monocular geometry estimation to facilitate multi-view reconstruction.

\paragraph{Leveraging Priors to Multi-View Geometry Estimation}
In multi-view geometry estimation, the fundamental matching assumption often breaks down within ill-posed regions (e.g., textureless areas, repetitive patterns). To tackle this challenge, existing approaches exploit structural priors to inject complementary geometric cues. For instance,
MonSter\cite{monster} improves reconstruction in ill-posed regions by explicitly fusing monocular depth predictions into multi-view optimization.
DepthSplat\cite{depthsplat}  demonstrates that integrating pre-trained monocular depth features into the multi-view matching pipeline substantially boosts novel view synthesis quality, achieving both architectural simplicity and state-of-the-art performance.
Inspired by these methods, we leverage enhanced robustness in monocular geometry estimation to facilitate geometric 3D reconstruction model.

\section{Method}

Given a pair of images $\{{\bm I}_{i}\}_{i=1}^2, ({\bm I}_i \in \mathbb{R}^{H \times W \times 3}$, where $H$ and $W$ are the image sizes) without pre-computed  camera extrinsics and intrinsics, our goal is to predict dense point maps of each frame $\{{\bm P}_{i}\}_{i=1}^2 ({\bm P}_i \in \mathbb{R}^{H \times W \times 3})$. The point map ${\bm P}_{i}$ associates each pixel $\bm y=(u,v)$ with its corresponding 3D scene point ${\bm P}_{i}(\bm y) \in \mathbb{R}^{3}$, expressed in the coordinate system of ${\bm I}_1$.
Note that this setting is different from the classical formulation of multi-view depth estimation\cite{schoenberger2016mvs, yao2018mvsnet} where all camera parameters are supposed to be provided as input. In our case, the camera parameters are intrinsically derived from our predicted pointmap representation.

Our framework comprises two complementary branches and a refinement module.
Following DUSt3R\cite{wang24dust3r:} architecture, the pairwise branch processes image pairs through feature matching to simultaneously establish cross-view correspondences and regress pairwise pointmaps. 
The monocular branch employs the state-of-the-art MoGe model~\cite{wang24moge:} to extract robust visual features and generate high-quality monocular pointmaps from single images.
The mono-guided refinement module integrates information from both branches through:
(i) Global Sim(3) alignment to establish a unified coordinate system for monocular pointmaps
(ii) An optimization procedure that refines pairwise pointmaps using monocular priors.
This design addresses inherent alignment noise while ensuring the final pointmaps maintain both multi-view consistency and single-view robustness. We provide detailed implementations of each component in subsequent sections.

\subsection{Pairwise Branch}
We adopt the DUSt3R framework~\cite{wang24dust3r:} for our pairwise processing module, which jointly extracts cross-view features and predicts consistent pointmaps.
The pipeline operates as follows: given a pair of images, a weight-sharing ViT\cite{dosovitskiy2021an} encoder independently processes each input image to extract initial features; to exchange information across different views, a dual-way decoder with 12 stacked Transformer blocks\cite{vaswani2017attention} (alternating self-attention and cross-attention layers) enables comprehensive information exchange between views, producing multi-view-aware features;
following, the dpt regression head\cite{ranftl21dpt} aggregates intermediate features from different decoder layers through layer-wise feature fusion, and then sends the fused feature to predict the pointmaps $\{{\bm P}_{i}^0\}_{i=1}^2$ of two frames in the coordinate system of $I_1$, along with the confidence of prediction $\{{\bm w}_{i}^0\}_{i=1}^2$.
We simply bilinearly interpolate the spatial resolution of the fused feature to restore spatial resolution of images and obtain the pairwise fetures $\{ \bm F_i \}_{i=1}^2$ (${\bm F}_i \in \mathbb{R}^{ {H} \times {W} \times C_{pair} }$, where $C_{pair}$ is the pairwise feature dimension).

The pairwise branch's ability to predict geometrically consistent pointmaps in a shared coordinate system provides the foundation for our subsequent refinement module, which performs deeper enhancement while preserving this cross-view consistency.

\begin{figure}[t!]
  \centering
  \includegraphics[width=0.6\columnwidth]{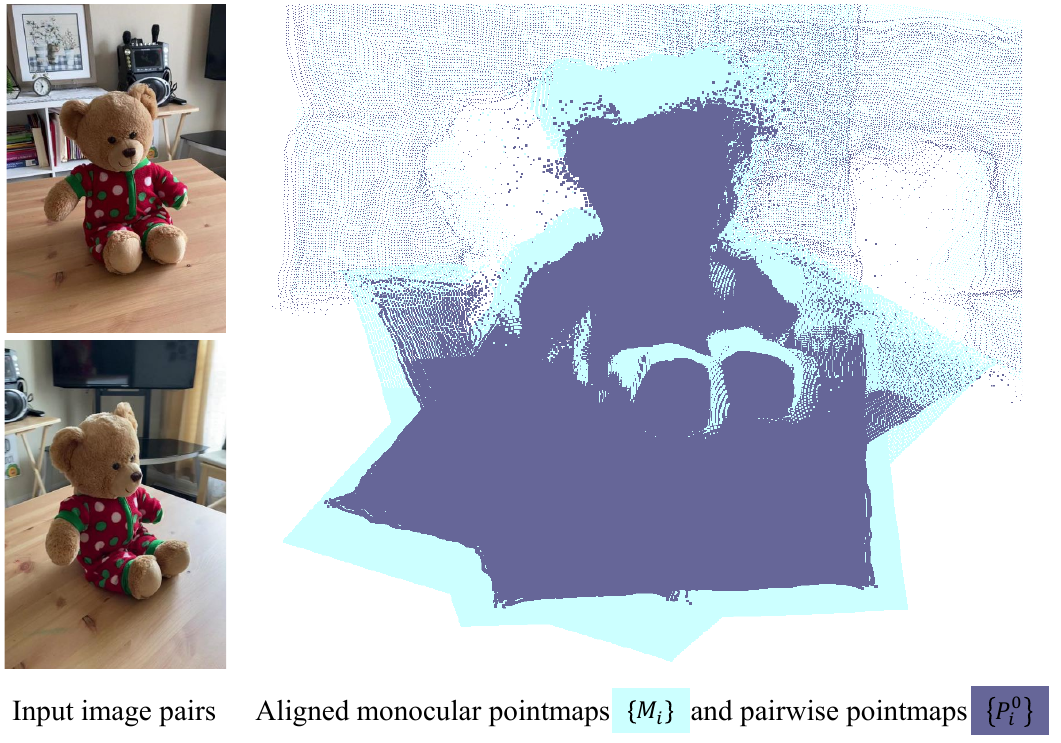}
  \caption{
    Comparision between aligned monocular pointmaps $\{M_{i}\}$ and pairwise pointmaps $\{P_i^0\}$.
    Although the monocular pointmap has undergone global alignment with the predictions from the pairwise branch, the aligned results still exhibit severe discrepancy.
  }
  \label{fig:mono-pair}
\end{figure}

\subsection{Monocular Branch}
While recent multi-view matching methods~\cite{wang24dust3r:,yang2025fast3r,wang2024spann3r,mast3r} have advanced geometry estimation, they remain fundamentally limited in challenging scenarios involving occlusions, texture-less regions, and reflective surfaces. To address these limitations, we augment the multi-view pipeline with monocular cues that provide robust scene understanding even in ambiguous conditions.
More specifically, we leverage the pre-trained monocular geometry backbone from the recent MoGe\cite{wang24moge:} model, selected for its proven effectiveness on diverse in-the-wild data. 
The monocular encoder is DINOv2\cite{oquab24dinov2:}, which has a patch size of 14 and outputs feature tokens for all input images in parallel.
A lightweight CNN-based head is used to extract affine-invariant pointmap $\{\hat{M}_{i}\}_{i=1}^2$ from these tokens.
Similar to pairwise branch, we bilinearly interpolate the feature from a intermediate stage of head to spatial resolution as the monocular feature $\{ \hat{\bm F}_i \}_{i=1}^2$ (${\bm F}_i \in \mathbb{R}^{ {H} \times {W} \times C_{mono} }$, where $C_{mono}$ is the monocular feature dimension).

The monocular branch demonstrates particular robustness in challenging scenarios where matching-based approaches typically fail, maintaining stable predictions even for texture-less surfaces and under complex lighting conditions\cite{wang24moge:}. This complementary strength motivates our design of fusing monocular predictions with multi-view reconstructions.

\subsection{Monocular Cues Guided Refinement}

The refinement module processes the initial monocular pointmaps  $\{\hat{M}_{i}\}_{i=1}^2$ and features $\{ \hat{F}_i \}_{i=1}^2$ through a two-stage procedure to enhance the pairwise pointmaps.
First, a global Sim(3) alignment is performed to compute the optimal similarity transformation that coarsely registers the monocular pointmaps with the pairwise predictions. This establishes an initial geometric consistency between the monocular and multi-view representations.
Subsequently, an iterative monocular-guided refinement process progressively optimizes the pairwise pointmaps by incorporating the aligned monocular priors as geometric constraints. 

\textbf{Global Sim(3) Alignment.} 
Global Sim(3) Alignment performs least squares optimization over a global scale $s_i^G$, shift $t_i^G$ and rotation $R_i^G$ to coarsely align each monocular pointmap with the corresponding pairwise pointmap: 
\begin{equation}
\begin{aligned}
s_i^G, t_i^G, R_i^G &= \arg\min \sum_{y} w_i^{0}(y) \left\| s_i^G \bigl( R_i^G \hat{M}_i(y) + t_i^G \bigr) - P_i^0(y) \right\|^2 \\
M_i &= s_i^G (R_i^G \hat{M}_{i} + t_i^G)
\end{aligned}
\label{equ:sim3_align}
\end{equation}
where $y \in H \times W$, $\{{M}_{i}\}_{i=1}^2$ denote the aligned monocular point maps, and $\{{w}_{i}^{0}\}_{i=1}^2$ acts as a confidence weight to exclude unreliable predictions, such as the sky, extreme depth ranges and hard regions.
Intuitively, this step converts the monocular pointmap coarsely aligned with the pairwise predictions, enabling effective refinement in the same space.

\begin{figure*}[t!]
  \centering
  \includegraphics[width=\textwidth]{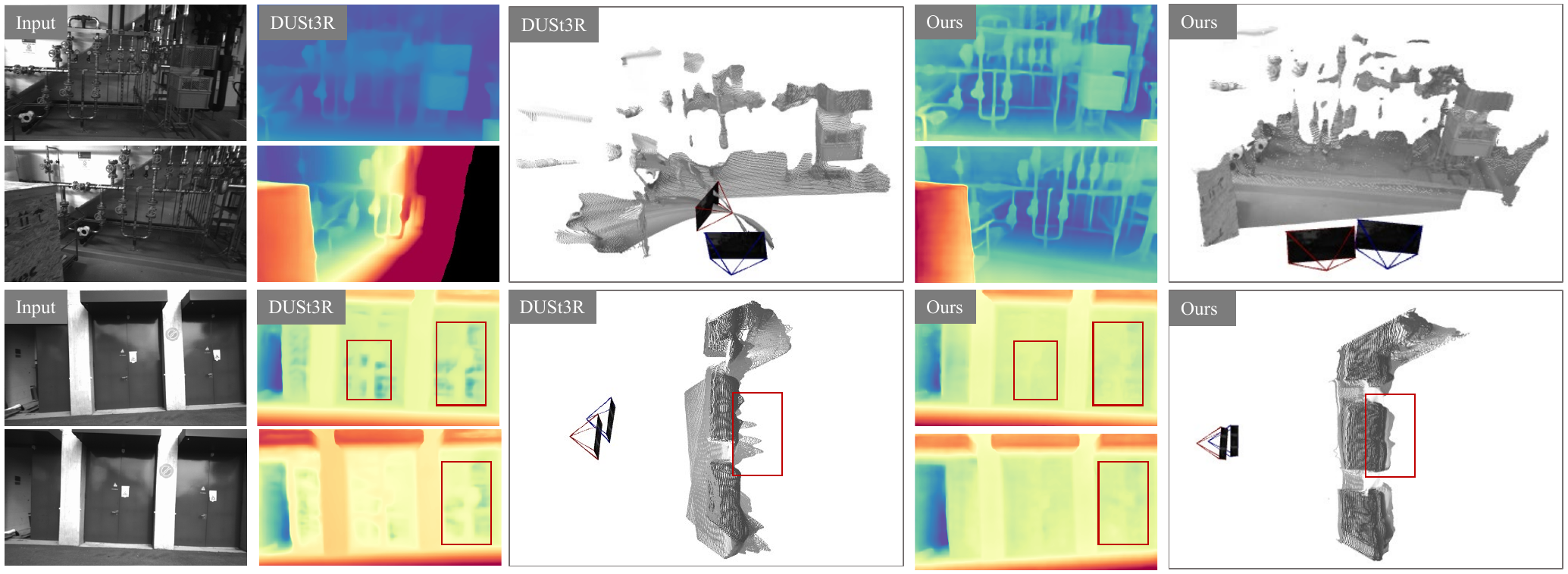}
  \caption{
  Qualitative comparison of our predicted depthmaps and 3D points to DUSt3R on in-the-wild captured images.
  Colored camera frustums illustrate the estimated camera poses.
  As shown in the top row, our method successfully predicts the thin tubular structure of metal pipes, while DUSt3R predicts a significantly distorted structure. 
  In the second row, our method robustly recovers the flat door structure from two images with repeated textures, while DUSt3R generates false depth discontinuities that violate the planar surface prior.
  }
  \label{fig:comp_pcd}
\end{figure*}

\textbf{Refinement.}
While a unified Sim(3) transformation provides coarse alignment, as shown in \cref{fig:mono-pair}, the aligned pointmaps $\{M_i\}$ exhibit residual misalignment with the pairwise pointmaps $\{P_i^0\}$. 
This misalignment arises from a combination of prediction noise in the point map estimation and inherent domain gaps between the monocular and pairwise estimation models.
To address this, we propose a monocular cues-guided refinement module that iteratively enhances pointmap accuracy.
First, we encode a condition feature by concatenating multi-modal inputs:
\begin{equation}
\begin{aligned}
x_{i}^{cond} = & \; \text{E}_{cond}([{M}_{i}, \hat{F}_{i}, {F}_{i} ,{w}_{i}, {I}_{i} ])\\
\end{aligned}
\end{equation}
where $\text{E}_{cond}$ is the a lightweight convolutional network for feature encoding. 
The condition feature $x_{i}^{cond}$ drives a ConvGRU-based updater to refine the hidden state $h_{m}^{i-1}$.
At step $j$:
\begin{equation}
\begin{aligned}
z^j = & \;\sigma(\text{Conv}([h_{M}^{j-1}, x_{S}^j], W_z) + c_k), \\
r^j = & \;\sigma(\text{Conv}([h_{M}^{j-1}, x_{S}^j], W_r) + c_r), \\
\Tilde{h}_{M}^j = & \,\tanh(\text{Conv}([r^j \odot h_{M}^{j-1}, x_{S}^j], W_h) + c_h), \\
h_{M}^j = & \;(1-z^j) \odot h_{M}^{j-1} + z^j \odot \Tilde{h}_{M}^j,
\end{aligned}
\label{equ:gru}
\end{equation}
where context features $c_k$, $c_r$, $c_h$ and initial state $h_{m}^{0}=tanh(\hat{F}_{i})$.
From the hidden state $h_{M}^{j}$, we decode a \emph{residual pointmap offset} $\Delta \bm{p}$ via convolutional layers to update the pairwise pointmaps:
\begin{equation}
\begin{aligned}
\bm{P}^{j+1}_{i} = & \; \bm{P}^j_{i} + \Delta \bm{p}.
\end{aligned}
\label{equ:update}
\end{equation}
After $N$ iterations, the final refined pointmap $\bm{P}^{N}_{i}$ is obtained
  
\subsection{Training Objective}
We use the 3D regression loss to supervise the output from pairwise branch and mono-guided refinement module. 
We denote the set of pointmaps from the $N$ iterations of the refinement module as $\{\bm{P}_{i}\}_{i=0}^{{N} -1}$ and follow \cite{teed20raft:, lipson21raft-stereo:} to exponentially increase the weights as the number of iterations increases. The total loss is defined as the sum of the pairwise branch loss $\mathcal{L}_{pair}$ and the refinement module loss $\mathcal{L}_{refine}$ as follows:
\begin{equation}
\begin{aligned}
     \mathcal{L}_{refine} = & \; \sum_{v=1}^{{N}}\sum_{i=1}^{{2}}\sum_{k=1}^{{H \times W}} \gamma^{{N}-v}  \left\Vert \frac{1}{\bm z}\bm{P}^{v}_{i,k}  - \frac{1}{\bm{\bar{z}}}\bm{\bar{P}}^{v}_{i,k} \right\Vert \\
    \mathcal{L}_{pair} = & \;  \sum_{i=1}^{{2}}\sum_{k=1}^{{H \times W}} {\bm w}^{0}_{k} \left\Vert \frac{1}{\bm{z}}\bm{P}^{0}_{i,k}  - \frac{1}{\bm{\bar{z}}}\bm{\bar{P}}^{0}_{i,k} \right\Vert - \alpha \log \bm{w}^{0}_{k}\\
\end{aligned}
\end{equation}
where $\gamma=0.9$, $\bm{\bar{P}}$ is the ground truth pointmaps, $\bm z$ and $\bm {\bar z}$ are the normalizing factor for predicted and ground-truth pointmaps respectively, and ${\bm w}^{v}_{0}$ is the confidence score for pixel $k$, which enable confidence-aware loss.

\begin{figure}[t!]
  \centering
  \includegraphics[width=0.6\linewidth]{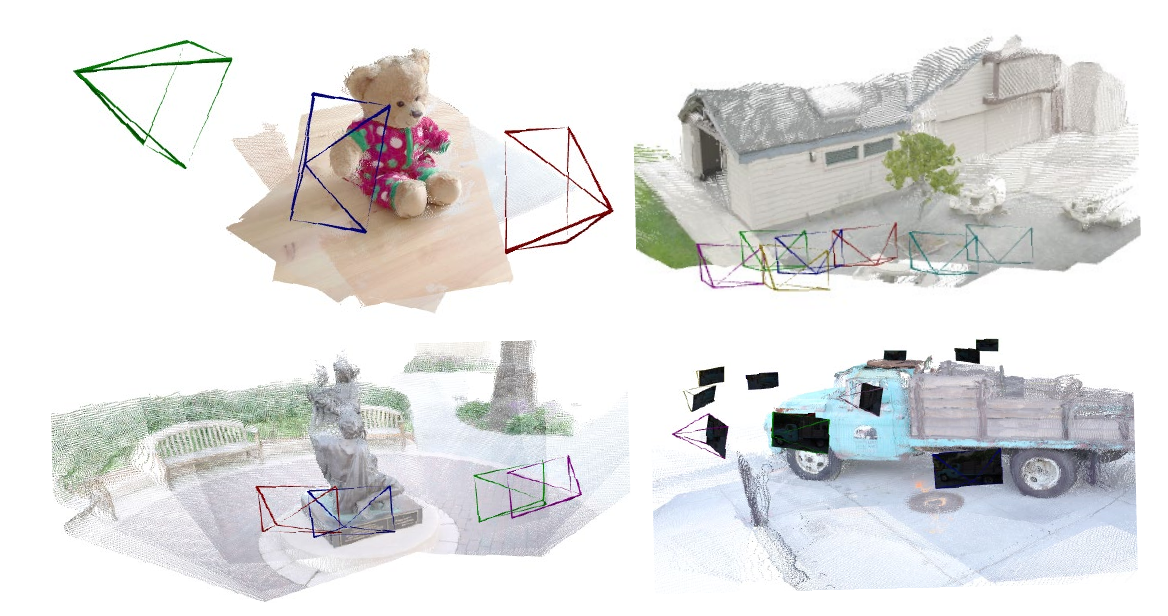}
  \caption{
    Qualitative examples of Mono3R’s output.
  }
  \label{fig:self_pcd}
\end{figure}

\section{Experimental Results}
In this section, we experiment with two representative 3D vision tasks, camera pose estimation and multi-view point cloud estimation in \cref{mvpose,mvpointmap}.
Then we present some detailed ablation study in \cref{ablation_study}.

\begin{table*}[htbp]
\setlength{\tabcolsep}{3pt}
\centering
\resizebox{\textwidth}{!}{%
\begin{tabular}{l l *{7}{c} *{4}{c}}
\toprule
\multirow{2}{*}{\textbf{Dataset}} & \multirow{2}{*}{\textbf{Method}} & \multicolumn{7}{c}{\textbf{Mutli-View Pose Estimation}} & \multicolumn{4}{c}{\textbf{Multi-View Stereo}} \\
\cmidrule(lr){3-9} \cmidrule(lr){10-13}
 & & $\text{mAA}_{30}\uparrow$ & $\text{RRA}_{5}\uparrow$ & $\text{RRA}_{10}\uparrow$ & $\text{RRA}_{15}\uparrow$ & $\text{RTA}_{5}\uparrow$ & $\text{RTA}_{10}\uparrow$ & $\text{RTA}_{15}\uparrow$ & $\text{Acc-Mean}\downarrow$ & $\text{Acc-Med}\downarrow$ & $\text{Comp-Mean}\downarrow$ & $\text{Comp-Med}\downarrow$ \\
\midrule
\multirow{5}{*}{\textbf{7Scenes}} 
& \textbf{DUSt3R} & 0.576 & 0.560 & 0.901 & 0.967 & 0.298 & 0.562 & 0.668 & 0.060 & 0.044 & 0.072 & 0.051 \\
& \textbf{Spann3R} & 0.187 & 0.138 & 0.292 & 0.419 & 0.068 & 0.158 & 0.248 & 0.081 & 0.060 & 0.143 & 0.110 \\
& \textbf{Fast3R} & 0.538 & 0.390 & 0.696 & 0.772 & 0.331 & 0.541 &  0.660 & 0.108 & 0.082 & 0.177 & 0.139 \\
& \textbf{Ours}  & \textbf{0.728} & \textbf{0.615} & \textbf{0.921} & \textbf{0.970} & \textbf{0.544} & \textbf{0.778} & \textbf{0.866} & \textbf{0.055} & \textbf{0.040} & \textbf{0.068} & \textbf{0.049} \\
& \textbf{$\Delta$} & \textbf{+0.152} & \textbf{+0.055} & \textbf{+0.020} & \textbf{+0.003} & \textbf{+0.246} & \textbf{+0.216} & \textbf{+0.198} & \textbf{-0.005} & \textbf{-0.004} & \textbf{-0.004} & \textbf{-0.002} \\
\cmidrule(r){1-13}
\multirow{5}{*}{\textbf{NRGBD}}
& \textbf{DUSt3R}    & 0.772 & 0.959 & 0.986 & 0.986 & 0.524 & 0.811 & 0.893 & 0.068 & 0.046 & 0.058 & 0.038 \\
& \textbf{Spann3R}   & 0.004 & 0.006 & 0.019 & 0.042 & 0.002 & 0.025 & 0.029 & 0.125 & 0.081 & 0.112 & 0.079 \\
& \textbf{Fast3R}    & 0.652 & 0.607 & 0.804 & 0.846 & 0.497 & 0.675 & 0.767 & 0.117 & 0.082 & 0.120 & 0.071 \\
& \textbf{Ours}  & \textbf{0.887} & \textbf{0.964} & \textbf{0.999} & \textbf{1.000} & \textbf{0.807} & \textbf{0.945} & \textbf{0.982} & \textbf{0.069} & \textbf{0.046} & \textbf{0.056} & \textbf{0.037} \\
& \textbf{$\Delta$} & \textbf{+0.115} & \textbf{+0.005} & \textbf{+0.013} & \textbf{+0.014} & \textbf{+0.283} & \textbf{+0.134} & \textbf{+0.089} & \textbf{+0.001} & \textbf{+0.000} & \textbf{-0.002} & \textbf{-0.001} \\
\bottomrule
\end{tabular}
}
\caption{Quantitative results on two indoor datasets, 7Scenes [67] and NRGBD [6] datasets, with absolute improvement ($\Delta$) between our method and DUSt3R. Positive values indicate improvements for metrics marked with $\uparrow$, while negative values indicate improvements for metrics marked with $\downarrow$.}
\label{tab:merged-benchmark-main-7scenes-nrgbd}
\end{table*}

{\textbf{Implementation Details.}}
The monocular encoder and decoder branch inherits the network architecture and pre-trained weights from MoGe\cite{wang24moge:}. In addition to the pointmap, MoGe's output includes a boolean mask indicating the validity of the pointmap. For instance, sky regions are often predicted as invalid areas. We set the predictions of these invalid regions to zero to prevent unreasonable coordinate values from affecting subsequent training. The head of MoGe employs a multi-layer ConvNet, from which we extract a 64-channel feature map $(C_{mono} = 64)$ at a specific layer and upsample it to match the image's height and width.
The pairwise encoder, decoder and dpt head inherits the network architecture and pre-trained weights from DUSt3R\cite{wang24dust3r:}. The decoder part consists of 12 blocks, each containing a self-attention layer and a cross-attention layer.
We extract a 128-channel feature map $(C_{pair} = 128)$ from the dpt head.
During training, we do not optimize the monocular branch but only the last two decoder blocks and head of the pairwise branch. This effectively balances efficiency and performance. We also discuss in the ablation study which parts to optimize and their impact on final performance and training time.
We use the Umeyama algorithm\cite{umeyama1991least} for Sim(3) alignment, with weights determined by the confidence predicted by the pairwise branch.
For the refinement stage, we adopt ConvGRU\cite{ballas2015delving} as the network architecture and set the number of iterations $N$ to 2. The ablation study examines how the number affects final performance and training time.
Due to computational resource constraints, we trained the model only at 224 resolution, which is sufficient to validate the effectiveness of our approach. 

{\textbf{Training and Testing Data.}}
We follow the training recipe from DUSt3R\cite{wang24dust3r:} to prepare training data. 
Namely, we use the provided pairs from a mix of datasets: MegaDepth\cite{li2018megadepth}, ARKitScenes\cite{dehghan2021arkitscenes}, Static Scenes 3D\cite{mayer16a-large}, BlendedMVS\cite{yao2020blendedmvs}, ScanNet++\cite{yeshwanthliu2023scannetpp}, Co3Dv2\cite{reizenstein21common} and Waymo\cite{waymo}. 
These datasets include indoor and outdoor/unbounded scenes, as well as real and synthetic ones.
The combination of our datasets is a subset to those of DUSt3R\cite{wang24dust3r:}, but comparable in size.
We train the network with a resolution of 224px for about 3 days on 8 V100 GPUs.
Our test datasets encompass object-level DTU\cite{dtudataset}, indoor-level 7Scenes\cite{7scenes} and NRGBD\cite{sturm2012benchmark}, as well as outdoor and unbounded ETH3D\cite{eth3d} and Tanks \& Temples\cite{tnt}. These test datasets are strictly disjoint from our training datasets. All testing is conducted at a fixed resolution of 224×224. Although the inference models provided by DUST3R and Fast3R were trained with mixed resolutions of 512 and 224, this does not affect the fairness of our evaluation.

{\textbf{Baselines.}}
DUSt3R\cite{wang24dust3r:} is the closest approach to ours, and competitive on visual odometry and reconstruction benchmarks. 
We additionally consider DUSt3R’s follow-up works Spann3R\cite{wang2024spann3r}, which seeks to replace DUSt3R’s expensive global alignment stage by sequentially processing frames with an external spatial memory. 
In our experiment, we use offline\footnote{Spann3r provides both offline and online modes. The offline mode achieves better performance for unordered image collections.} mode of Spann3R.
We also compare with Fast3R\cite{yang2025fast3r} that serves as a multi-view generalization to DUSt3R that achieves scalable 3D reconstruction by processing many views in parallel.

\begin{figure*}[t!]
  \centering
  \includegraphics[width=\textwidth]{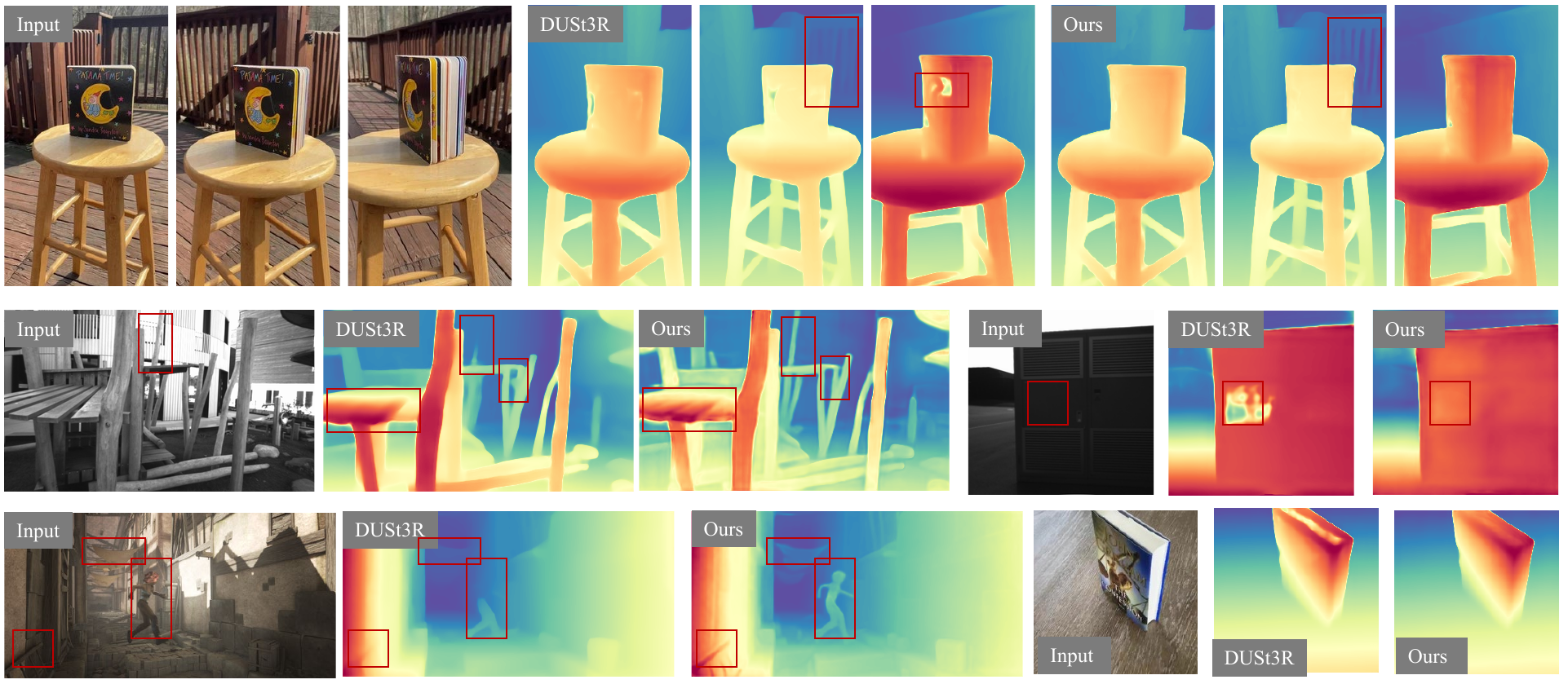}
  \caption{Additional visualizations of depth map estimation. Compared to DUSt3R, our prediction is high-quality when given challenging examples with repeated textures and thin structure.}
  \label{fig:comp_depth}
\end{figure*}

\subsection{Multi-View Pose Estimation}
\label{mvpose}

For camera pose evaluation, we employ scale-invariant metrics to assess relative rotation and translation accuracy (RRA and RTA, respectively). These metrics are evaluated at task-specific thresholds (5°, 10°, and 15°), along with the mean Average Accuracy (mAA) - defined as the area under the accuracy curve for angular differences at min(RRA, RTA). Note that higher values indicate better performance for all metrics.
We evaluate Mono3R for the task of multi-view pose prediction on the 7scenes\cite{7scenes}, DTU\cite{dtudataset}, NeuralRGBD\cite{sturm2012benchmark} and Tanks \& Temples\cite{tnt} datasets.
For each sequence, we pick 10 random frames and report the results in \cref{tab:benchmark-main-DTU,tab:merged-benchmark-main-7scenes-nrgbd,tab:benchmark-main-eth3d,tab:benchmark-main-tnt}.
Our results show that our model consistently outperforms competing methods in all metrics.
This validates the superior generalization of our method.

\begin{table}[t!]
\centering
\begin{tabular}{lcccc}
\toprule
\textbf{Metric} & \textbf{DUSt3R} & \textbf{Spann3R} & \textbf{Fast3R} & \textbf{Ours} \\
\midrule
$\text{mAA}_{30}\uparrow$ & 0.742 & 0.469 & 0.692 & \textbf{0.776} \\
$\text{RRA}_{5}\uparrow$ & 0.902 & 0.343 & 0.505 & \textbf{0.909} \\
$\text{RRA}_{10}\uparrow$ & 0.975 & 0.527 & 0.760 & \textbf{0.990} \\
$\text{RRA}_{15}\uparrow$ & 0.990 & 0.629 & 0.863 & \textbf{1.000} \\
$\text{RTA}_{5}\uparrow$ & 0.465 & 0.283 & 0.465 & \textbf{0.522} \\
$\text{RTA}_{10}\uparrow$ & 0.742 & 0.465 & 0.741 & \textbf{0.785} \\
$\text{RTA}_{15}\uparrow$ & 0.857 & 0.560 & 0.835 & \textbf{0.894} \\
\midrule
$\text{Acc-Mean}\downarrow$ & 3.500 & 4.379 & 6.347 & \textbf{3.440} \\
$\text{Acc-Med}\downarrow$ & 2.560 & 3.078 & 4.343 & \textbf{2.559} \\
$\text{Comp-Mean}\downarrow$ & 3.623 & 4.064 & 6.761 & \textbf{3.433} \\
$\text{Comp-Med}\downarrow$ & 2.407 & 2.723 & 4.659 & \textbf{2.274} \\
\bottomrule
\end{tabular}
\vspace{1em}
\caption{Quantitative results on object-level DTU\cite{dtudataset} dataset. Our method consistently demonstrates superior performance compared to all baseline approaches across all evaluation metrics.}
\label{tab:benchmark-main-DTU}
\end{table}

\begin{table}[t!]
\centering
\begin{tabular}{lcccc}
\toprule
\textbf{Metric} & \textbf{DUSt3R} & \textbf{Spann3R} & \textbf{Fast3R} & \textbf{Ours} \\
\midrule
$\text{mAA}_{30}\uparrow$     & \textbf{0.520}      & 0.015      & 0.458 & 0.511 \\
$\text{RRA}_5\uparrow$        & 0.690      & 0.190      & 0.506 & \textbf{0.800} \\
$\text{RRA}_{10}\uparrow$     & 0.813      & 0.278      & 0.641 & \textbf{0.876} \\
$\text{RRA}_{15}\uparrow$     & 0.869      & 0.344      & 0.707 & \textbf{0.911} \\
$\text{RTA}_5\uparrow$        & \textbf{0.307}      & 0.006      & 0.240 & 0.265 \\
$\text{RTA}_{10}\uparrow$     & \textbf{0.488}      & 0.019      & 0.427 & 0.462 \\
$\text{RTA}_{15}\uparrow$     & \textbf{0.607}      & 0.028      & 0.581 & 0.585 \\
\bottomrule
\end{tabular}
\vspace{1em}
\caption{Quantitative results on Outdoor ETH3D\cite{eth3d} dataset. Our method demonstrates significant superiority in RRA metrics, achieves comparable performance in RTA, and maintains competitive overall precision.}
\label{tab:benchmark-main-eth3d}
\vspace{-1em}
\end{table}

\subsection{Multi-View Point Cloud Estimation}
\label{mvpointmap}

When evaluating point map, the performance is evaluated in terms of accuracy, which is the smallest Euclidean distance to the ground-truth, and completeness as the smallest distance to the reconstructed shape, with the overall average.
We compare the accuracy of our predicted point cloud to DUSt3R\cite{wang24dust3r:}, Spann3R\cite{wang2024spann3r} and Fast3R\cite{yang2025fast3r} on the DTU\cite{dtudataset}, 7Scenes\cite{7scenes} and NRGBD\cite{sturm2012benchmark} dataset, covering overall 50+ diverse scenes.
For each scene, we randomly sample 10 frames.
The predicted point cloud is aligned to the ground truth using the Umeyama algorithm\cite{umeyama1991least}.
We report the mean and median of accuracy and completeness for point map estimation.
As shown in \cref{tab:benchmark-main-DTU,tab:merged-benchmark-main-7scenes-nrgbd}, our method outperforms other approaches consistently.
Meanwhile, We present a qualitative comparison with DUSt3R on in-the-wild scenes in \cref{fig:comp_pcd} and further examples in \cref{fig:comp_depth}. Mono3R outputs high-quality predictions and generalizes well, excelling on challenging out-of-domain examples, such as scenes with repeating or homogeneous textures.

\begin{table}[t!]
\centering
\begin{tabular}{lcccc}
\toprule
\textbf{Metric} & \textbf{DUSt3R} & \textbf{Spann3R} & \textbf{Fast3R} & \textbf{Ours} \\
\midrule
$\text{mAA}_{30}\uparrow$     & 0.800      & 0.000      & 0.766 & \textbf{0.859} \\
$\text{RRA}_5\uparrow$        & 0.750      & 0.058      & 0.720 & \textbf{0.903} \\
$\text{RRA}_{10}\uparrow$     & 0.948      & 0.122      & 0.884 & \textbf{0.979} \\
$\text{RRA}_{15}\uparrow$     & 0.988      & 0.213      & 0.963 & \textbf{0.999} \\
$\text{RTA}_5\uparrow$        & 0.680      & 0.000      & 0.595 & \textbf{0.779} \\
$\text{RTA}_{10}\uparrow$     & 0.892      & 0.000      & 0.843 & \textbf{0.927} \\
$\text{RTA}_{15}\uparrow$     & 0.936      & 0.000      & 0.906 & \textbf{0.963} \\
\bottomrule
\end{tabular}
\vspace{1em}
\caption{Quantitative results on Tanks \& Temples\cite{tnt} dataset. Our method demonstrates significant superiority in RRA metrics, achieves the best performance in RTA, and maintains the overall precision lead with the highest mAA score.}
\label{tab:benchmark-main-tnt}
\end{table}

\subsection{Ablation Study}
\label{ablation_study}

To validate the effectiveness of individual components in Mono3R, we conduct comprehensive ablation studies. All experiments maintain identical hyperparameter settings.

\begin{figure*}[t!] 
  \centering
  \includegraphics[width=\linewidth]{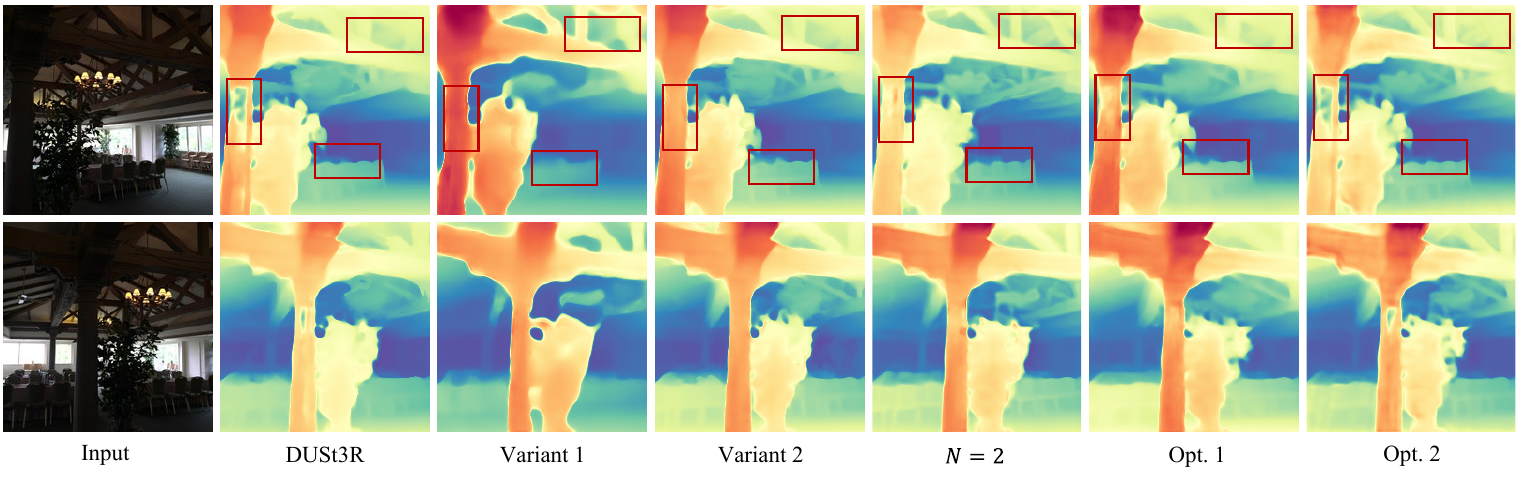} 
  \caption{Depth map visualization comparison between DUSt3R, two ablation variants, and our proposed method. Our approach demonstrates superior performance in preserving fine geometric details.}
  \label{fig:comp_ablation}
\end{figure*}

\subsubsection{Model variants}
In addition to our proposed mono-guided refinement module, we investigated two alternative strategies for fusing monocular geometric priors to comprehensively analyze their impact on model performance. The experimental designs are detailed below: 
\textbf{(1) Pointmap and Image Concatenation}: The pointmap generated by the monocular geometric model (MoGe) is concatenated with the original RGB image along the channel dimension (increasing channels from 3 to 6) as joint input. This requires retraining all downstream modules due to altered input dimensions. We initialized the decoder and head with DUST3R's pretrained weights, but the overall training time increased significantly due to more optimizable parameters.
\textbf{(2) Decoder-Level Feature Fusion}. In this variant, monocular prior features are dynamically fused with DUSt3R's decoded features during the decoder stage. 
We select 4 critical blocks (layers 0, 4, 8, 11) and insert lightweight feature fusion modules at their inputs. Here, the encoder remains frozen, while only the decoder and head are fine-tuned to adapt to prior information.
The corresponding results are presented in rows 2-3 of \cref{tab:ablation_table}, Variant 1 and Variant 
2 respectively. 
As can be observed, both alternative schemes lead to significant performance degradation, which further demonstrates the necessity of the mono-guided refinement module.
The qualitative results are shown in \cref{fig:comp_ablation}.

\begin{table}[t!]
\centering
\begin{tabular}{lccccc}
\toprule
\textbf{Method} & \textbf{$\text{mAA}_{30}\uparrow$} & \textbf{$\text{RRA}_{15}\uparrow$} & \textbf{$\text{RTA}_{15}\uparrow$} & \textbf{Acc$\downarrow$} & \textbf{Comp$\downarrow$} \\
\midrule
Variant 1 & 0.580 & 0.951 & 0.682 & 4.394 & 3.713 \\
Variant 2 & 0.687 & 0.990 & 0.788 & 3.437 & 3.523 \\
\midrule
$N=1$ & 0.724 & 0.995 & 0.834 & 3.869 & 3.535 \\
$N=2$ & 0.714 & 0.995 & 0.833 & 3.858 & 3.507 \\
$N=3$ & 0.730 & 0.995 & 0.840 & 3.856 & 3.510 \\
$N=4$ & $\textbf{0.732}$ & 0.995 & $\textbf{0.864}$ & $\textbf{3.921}$ & $\textbf{3.471}$ \\
\midrule
Opt. 1 & 0.693 & 0.990 & 0.819 & 4.345 & 3.402 \\
Opt. 2 & 0.641 & 0.941 & 0.765 & 4.925 & 3.630 \\
\bottomrule
\end{tabular}
\vspace{1em}
\caption{Ablation study on DTU\cite{dtudataset} dataset. The method \textit{$N=2$} is same as the setting of main experiment, except in number of training pairs.}
\label{tab:ablation_table}
\end{table}

\subsubsection{Numer of refinement iterations}
In this experiment, we adjusted the parameter $N$ from 1 to 4. Theoretically, as $N$ increases, the monocular-guided refinement module can perform more iterative refinements to further improve accuracy. 
We conducted studies on the DTU\cite{dtudataset} dataset, with experimental results shown in rows 4-6 of \cref{tab:ablation_table}.
The results demonstrate that the overall pose accuracy ($\text{mAA}_{30}$) exhibits an upward trend with increasing $N$, primarily driven by improvements in translation accuracy $\text{RTA}_{15}$ while rotation accuracy $\text{RRA}_{15}$ remains largely unchanged. 
Regarding point cloud accuracy, the enhancement is mainly reflected in the completeness metric.

\subsubsection{Optimization Strategies}
In our experiments, in addition to the essential optimization of the refinement module, we kept the monocular branch frozen without any optimization, and for the pairwise branch, only the last two decoder blocks and head were optimized. Here, we validated two alternative experimental schemes with fewer optimized parameters: 
\textbf{(1) optimizing only the head of the pairwise branch and refinement module} and \textbf{(2) optimizing solely the refinement module}.
The experimental results are shown in rows 7-8 of \cref{tab:ablation_table} and \cref{fig:comp_ablation}, Opt. 1 and Opt. 2 respectively.
Both schemes led to significant performance degradation in terms of both pose accuracy and point cloud accuracy, as demonstrated by the experimental results.

\section{Conclusion}
With the release of large-scale high-quality 3D datasets, the field of 3D reconstruction has finally witnessed a significant paradigm shift - transitioning from per-scene reconstruction to data-driven generalizable inference models. This transition is particularly exemplified by approaches like DUSt3R. However, due to their matching-based principle, such models still struggle with ill-posed regions having limited matching cues, including occlusions, textureless areas, and repetitive/thin structures.
To address this limitation, we present Mono3R, a novel framework built upon DUSt3R that significantly enhances reconstruction robustness and performance in challenging regions. By effectively injecting monocular geometric priors into DUSt3R's pipeline, Mono3R demonstrates superior performance across multiple public indoor and outdoor datasets.
We hope that Mono3R can make valuable contributions to the 3D vision community.



\bibliographystyle{unsrt}  
\bibliography{main}

\end{document}